\documentclass[letterpaper, 10 pt, conference]{ieeeconf} 
\IEEEoverridecommandlockouts 
\usepackage{graphicx}
\usepackage{amssymb}
\usepackage{amsmath}
\usepackage{threeparttable}
\usepackage{algorithm}
\usepackage{algorithmicx}
\usepackage{algpseudocode}
\usepackage{cite}
\usepackage{url}
\usepackage{color}
\usepackage{bm}
\usepackage{footnote}

\makeatletter
\let\NAT@parse\undefined

\newcommand{\Rmnum}[1]{\expandafter\@slowromancap\romannumeral #1@}
\makeatother

\newcommand{\Gcal}{\mathcal{G}}
\newcommand{\Hcal}{\mathcal{H}}
\newcommand{\Ncal}{\mathcal{N}}
\newcommand{\Rcal}{\mathcal{R}}
\newcommand{\Jcal}{\mathcal{J}}
\newcommand{\Mcal}{\mathcal{M}}

\usepackage[colorlinks, linkcolor=black, anchorcolor=black, citecolor=black]{hyperref}

\title
{\LARGE \bf
Learning to Collide: \\
An Adaptive Safety-Critical Scenarios Generating Method}

\author{Wenhao Ding$^{1}$, Baiming Chen$^{2}$, Minjun Xu$^{1}$ and Ding Zhao$^{1}$
\thanks{
$^{1}$Wenhao Ding, Baiming Chen, Minjun Xu and Ding Zhao are with the Department of Mechanical Engineering, Carnegie Mellon University, Pittsburgh, PA 15213, USA. {\tt\small wenhaod@andrew.cmu.edu}, {\tt\small minjunxu@andrew.cmu.edu}, {\tt\small dingzhao@cmu.edu}
}%
\thanks{
$^{2}$Baiming Chen is with the Department of Automotive Engineering, Tsinghua University, Beijing, 100084, China. {\tt\small cbm@mails.tsinghua.edu.cn}
}%
}

\begin{document}
\maketitle

\begin{abstract}
Long-tail and rare event problems become crucial when autonomous driving algorithms are applied in the real world.
For the purpose of evaluating systems in challenging settings, we propose a generative framework to create safety-critical scenarios for evaluating specific task algorithms. 
We first represent the traffic scenarios with a series of autoregressive building blocks and generate diverse scenarios by sampling from the joint distribution of these blocks.
We then train the generative model as an agent (or a generator) to search the risky scenario parameters for a given driving algorithm. 
We treat the driving algorithm as an environment that returns high reward to the agent when a risky scenario is generated. The whole process is optimized by policy gradient reinforcement learning method. 
Through the experiments conducted on several scenarios in the simulation, we demonstrate that the proposed framework generates safety-critical scenarios more efficiently than grid search or human design methods. Another advantage of this method is its adaptiveness to the routes and parameters.

\end{abstract}

\section{Introduction}

Nowadays, the performance of most perception and prediction algorithms are quite sensitive to the imbalance of the training data (also known as long-tail problem \cite{38}\cite{28}). 
Rare events are often difficult to collect and easily neglected in the huge data flow, which greatly challenges the real-world application of robots especially in safety-critical domains, e.g., autonomous driving. 

In the industry, companies usually resort to simulations to reproduce the safety-critical scenarios collected during their test driving. 
One brute-force method is creating risky scenarios by adjusting all variables in one scenario with the grid search to create similar risky scenarios, which is time and labor intensive.
An alternative method known as worst-case evaluation \cite{36} is proposed to search the worst cases for evaluating controllers in the vehicle field.
Although some cases excavated by worst-case evaluation may be useful, some extremely risky scenarios are almost impossible to appear in the real world. Evaluating algorithms in these scenarios may not represent real-world deployment.
Therefore, other previous works concentrate on generating risky and reasonable scenarios \cite{18}\cite{39} with importance sampling. 
These works show the possibility of modeling the scenarios with probability distributions and more efficient sampling from them.

With the recent popularity of deep generative models \cite{20} \cite{21}, another promising way is directly generating safety-critical scenarios instead of sampling from existing data. 
The advantage of the generative model is that more diverse scenarios, even the open-world scenarios that do not exist in the collected data, could be created.
Unfortunately, most prevalent deep generative models are not designed for generating rare events. 
These models generate random samples that are similar to the given dataset with no explicit generating process elaborated \cite{15}, which is not in line with our goal that generating rare and risky events with extremely low probability. 

Another difficulty of generating risk scenarios is finding the proper representation, since scenarios consisting of statistic and dynamic objects, route, and maps, are hard to model.
A low-dimensional and easy-to-sample representation would dramatically increase the efficiency of generating new scenarios.

\begin{figure}[t]
\centering
\includegraphics[width=8.5cm]{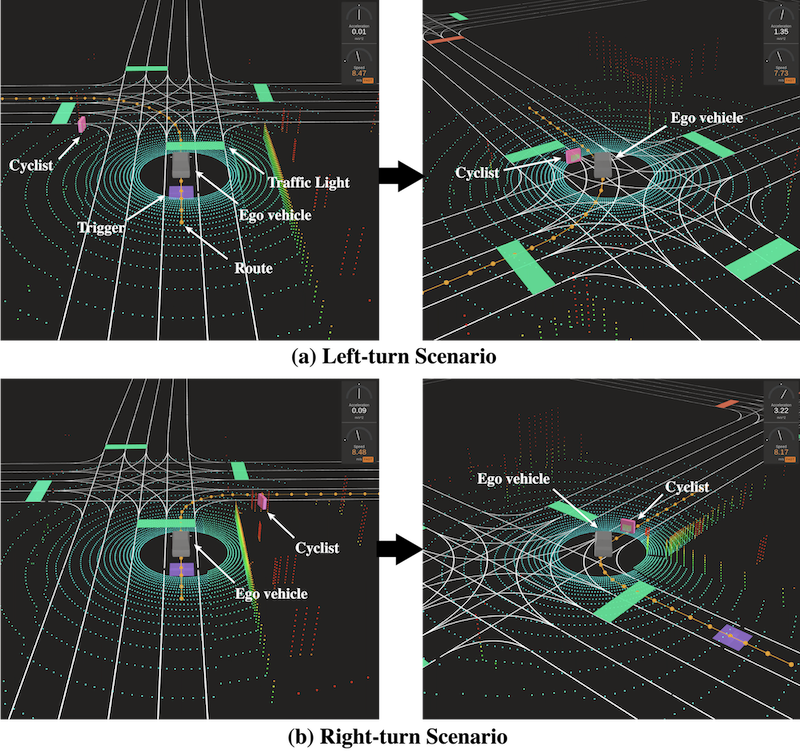}
\caption{Two generated safety-critical scenarios are displayed using our CarlaViz visualization tool (\href{https://github.com/wx9698/carlaviz}{Link}). In these scenarios, the ego vehicle is passing through an intersection and encounters a cyclist after taking a left or right turn. The left column shows the initial state and the right column shows the pre-crash state.} 
\label{overview}
\end{figure}

To address such challenges, we propose combining the task algorithm and the data generation model in this paper. Two generated scenarios examples are shown in Fig.~\ref{overview}.
Firstly, we use a factorized graphic model to represent the traffic scenario, which is inspired by \cite{41} and \cite{37}. 
The motivation is to integrate human knowledge to create scenarios by factorizing the scenarios into (independent) conditional probabilities.  We refer to the probabilities as the building blocks in this paper since they could be shared among different scenarios.
Using this graphic model allows us to generate new scenarios by sampling from the distributions according to the dependence relationship.
These building blocks vary the initial parameters to cover a wide range of scenarios, which was shown to be an effective method in modeling transportation  \cite{41}. 
After representing the scenario, we consider the generative model as an agent (or a generator) and regard one specific task algorithm as the environment. 
The parameters (e.g. target speed) and the route of the task algorithm will be used as input states for the agent. The riskier scenario the agent generates, the higher reward it receives. 
We sample the input states from a uniform distribution during the model’s training.
Therefore, given a specific task algorithm, our proposed framework adaptively generates safety-critical scenarios for different routes and input parameters, even for the settings that are not seen in the training stage.

In summary, this paper makes three contributions to the autonomous driving safety literature:
\begin{itemize}
	\item We develop a framework to generate traffic scenarios by sampling from the joint distributions of autoregressive building blocks.
	\item We design an algorithm that uses the task algorithm to guide the generation module for safety-critical scenario generation.
	\item The generated scenarios could be used to evaluate the safety of driving algorithms. The method could also be extended to create test scenarios for other types of robots.
\end{itemize}

\section{Related Work}

\subsection{Imbalanced-class and long-tail problem}

In the classification area, the long-tail problem has been studied for a long history. 
The models tend to perform poorly when they are trained on an imbalanced dataset where one class contains fewer samples than others. 
To tackle this problem, several kinds of methods have been proposed. 
The first kind is the re-weighting method proposed in \cite{27}, which adaptively modifies the loss function to force the model to focus more on the rare class. 
The second kind is the over-sampling method proposed in \cite{26}, which tries to create more samples of the rare class and trains the model on an artificially balanced dataset. 
The third one is a meta-learning related model \cite{28}, which improves the adaptive capability of models by leveraging the information among different domains. 
All these ideas are intuitive but efficient, thus they provide promising insights for dealing with tasks that are related to rare events.

Autonomous driving is a crucial area restricted by long-tail and open-world problems since autonomous vehicles (AV) in the real world frequently encounter new scenarios.
In the literature of adversarial attack \cite{45}\cite{46}, the worst cases are excavated to evaluate an algorithm's ability to handle high-risk scenarios.
There is a low probability, however, of these extreme scenarios appearing in the real world.
For example, we control a car to rush towards the ego vehicle according to the route and velocity of the latter, which is likely to cause a car crash.

To avoid creating unrealistic scenarios, \cite{18} and \cite{19} propose to use importance sampling to accelerate evaluation. 
They consider not only the rareness but also the rationality when sampling from the distribution of existing scenarios. 
Their way of thinking from the perspective of probability is very enlightening to us.

\subsection{Data generation with generative models}

The generative model is another choice to provide risky and rare scenarios. 
Recently, with the upsurge of deep learning, lots of generative models are prevalent again. 
For instance, Generative Adversarial Nets (GAN) \cite{20} and Variational Auto-encoder (VAE) \cite{21} are good for generating images, and Flow-based model \cite{22} and Autoregressive (AR) Model \cite{23} are usually used for generating sequences. 
The basic principle behind generative models is the assumption that there exists a data generation mechanism, with which we can generate samples following preferred distributions.
The objective function of these models can be categorized as a maximum likelihood estimation. 
In GAN models, the implicit likelihood is the decision of the discriminator, while in VAE models, the variational inference is leveraged to calculate the explicit likelihood.

Follow this line, prior works have attempted to generate scenarios with GAN or VAE involved. 
\cite{17} modifies the last layer of a generative adversarial imitation learning model to create some risky scenarios. 
\cite{15} projects the high-dimensional traffic trajectory data into a low-dimensional latent space with VAE and then samples the latent variables to regenerate scenarios with more diversity. 
In \cite{16}, the authors mix the collision data and the safety data in a disentangled latent space to generate intermediate near-miss scenarios.

Although these methods show the possibility of generating scenarios with generative models, two problems are still remained to be solved. 
The first one is how to sample rare events from these learned generative models since the rare events usually have low probability. 
The second one is how to prove the generated scenarios are indeed safety-critical and reasonable for the AV.

\begin{figure}[t]
\centering
\includegraphics[width=8.5cm]{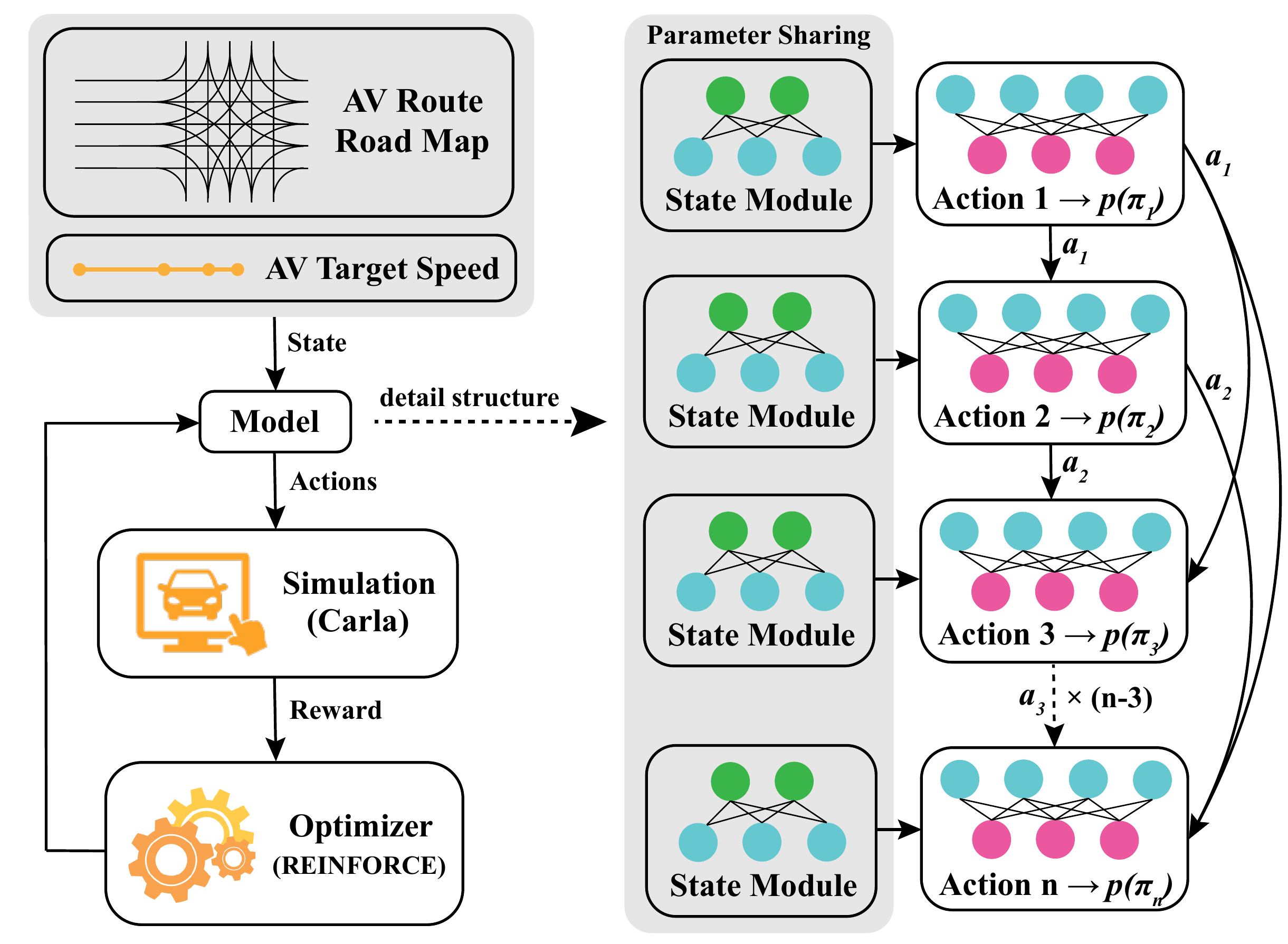}
\caption{The proposed framework (left) and the structural details (right). Our model consists of two parts: the state module and the action module, both of which are implemented with linear layers.}
\label{structure}
\end{figure}

\subsection{Task-guided data generation}

To obtain rare or safety-critical data, we do not have to model the entire data distribution.
The only thing we care about is the distribution of the rare event. 
Therefore, we combine the task and the generative procedure and use the task algorithm to guide the generative model so that it only learns the rare event distribution. 
Previously, this idea has been explored in many fields. 
Before combined with tasks, traditional data augmentation \cite{7}\cite{8} or domain randomization (DR) \cite{5}\cite{6} methods suffered from the problem of inefficient sampling, and task-guided methods such as those proposed in \cite{10}\cite{11}\cite{12}\cite{13} and \cite{29} significantly improved the performance. 
\cite{10} searches the data augmentation policy according to the accuracy of a recognition model (task algorithm) and they use adversarial training to optimize both the recognition model and policy selection model simultaneously. 
For domain randomization methods, \cite{11} and \cite{13} also propose to generate new domains by decreasing the accuracy of the task model. 
They use the task algorithm as guidance to generate adversarial domains. Although these works do not focus on rare events, the studies motivate our use of the evaluation procedure to generate safety-critical scenarios.
Adaptive Stress Testing \cite{42}\cite{43}\cite{44} is a kind of methods that share similar idea with ours. They also use the Mento Carlo Tree Search or Deep Reinforcement Learning method to find failure cases in airborne and traffic scenarios.

\begin{figure}[t]
\centering
\includegraphics[width=8cm]{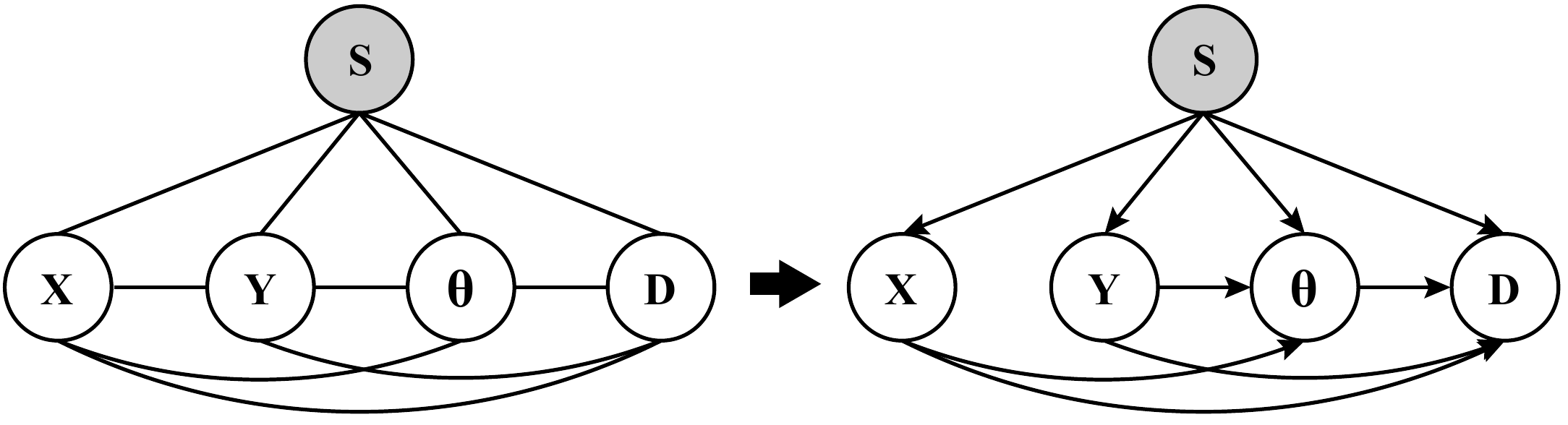}
\caption{One example of converting an undirected graphic model to a directed one with human knowledge. After the conversion, we are able to factorize the graph using conditional probabilities.}
\label{graph}
\end{figure}

\begin{figure}[t]
\centering
\includegraphics[width=7cm]{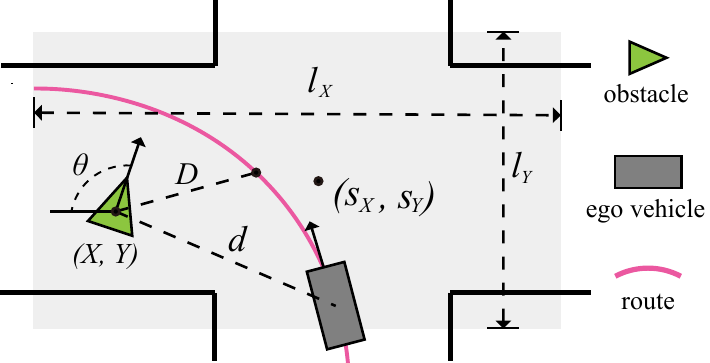}
\caption{Explanation of the nodes in Fig~.\ref{graph}. The scale and shift parameters in (\ref{rescale}) are also shown in this figure.}
\label{explain}
\end{figure}

\section{Method}

We have shown our proposed framework in Fig.~\ref{structure}. We firstly introduce how we represent the scenario generation process (right part of Fig.~\ref{structure}), then describe the pipeline of our model and the training procedures (left part of Fig.~\ref{structure}).

\subsection{Scenario representation}
\label{scenario04}

Most of the traffic scenarios could be divided into several building blocks, e.g., location, orientation, and velocity of traffic participants \cite{14}. 
Mathematically, we use a probability graphic model to represent a scenario where each node represents a building block. 

The most intuitive way to model the scenarios is using the undirected graph like \cite{41} when we have no information about the elements in the scenario.
We use a joint probability $P(\pi_1, \pi_2,...\pi_n)$ to represent one scenario, where $\pi_i$ is the distributions of setting parameters, such as the orientation or velocity of a cyclist. Then, we represent the scenario with an undirected graph $\Gcal$:
\begin{equation}
	P(\pi_1,...,\pi_n) = \frac{1}{Z} \prod_{v\in V} \Psi(\bm{\pi}_v)
\end{equation}
where $V$ indicates all nodes in the graph $\Gcal$ and $\Psi(\cdot)$ is the potential function to describe the correlation associated with the nodes of $\Gcal$. $Z$ is the partition function ensuring that the representation satisfies the requirement of probability. The drawbacks of this representation are two-fold: (1) $Z$ is usually intractable because it requires the integration of all latent variables, (2) $\Psi(\bm{\pi}_c)$ is difficult to define. Therefore, we simplify this representation to a directed graph $\Hcal$ by introducing human knowledge:
\begin{equation}
	P(\pi_1,...\pi_n) = \prod_{i\in V} P(\pi_i) \prod_{j\in V} P(\pi_{j}|\pi_{pa(j)}) 
\label{eq1}
\end{equation} 
where $pa(j)$ is the set of parents of $j$.  
The first production in (\ref{eq1}) represents the independent building blocks, while the second production usually represents the blocks with an autoregressive structure. Autoregressive structure means several nodes $\{v_0,...,v_g \} \in V$ form a specific group with the following relation:
\begin{equation}
	P(\pi_{j}| \pi_{0},...,\pi_{j-1}), \ \ for\ j\in \{1,...,g\}
\end{equation}
Human knowledge of traffic scenarios could help with factorizing the graph and design a Bayesian Network that represent the relation among all blocks.

A factorization example is shown in Fig.~\ref{graph}. 
In this scenario, we want to spawn one cyclist and move it when the AV reaches the trigger distance to the cyclist. 
We define four nodes to represent the scenario: X spawn position $({X})$, Y spawn position $({Y})$, orientation $({\Theta})$ and trigger distance $({D})$. 
$({S})$ represents the input states such as the route that the AV will follow and its target speed. 
The factorized result shown on the right of Fig.~\ref{graph} is the same as:
\begin{equation}
\begin{split}
	&P(X, Y, \Theta, D|S) \\
	=& P(X|S)P(Y|S)P(\Theta|S,X,Y)P(D|S,X,Y,\Theta)
\end{split}
\end{equation}

\begin{algorithm}[t]
    \centering
    \caption{Training Procedure of Proposed Framework}
    \label{alg1}
  	\begin{algorithmic}[1]
  	\State Initiate max epoch number $E$
	\State Initiate rewards $\bm{r_{b}}$, $\bm{r_{p}}$
	\State Build the model $\Mcal$ with (\ref{eq1}) and initiate parameters $\phi$
	\For{$e$ $\leftarrow$ 1 to $E$}
		\For{$i$ $\leftarrow 1$ to $N$}
    		\State $\bm{\xi}^{(i)}, \bm{\eta}^{(i)} \sim S$\ \Comment{sample states}
			\State $\bm{a}^{(i)}$ $\leftarrow$ $\Mcal$ ($\bm{\xi}^{(i)}$, $\bm{\eta}^{(i)}$; $\phi$)\ \Comment{sample actions}
			\State $\rho_{AV}$, $\rho_o$ $\leftarrow$ Simulator ($\bm{\xi}^{(i)}$, $\bm{\eta}^{(i)}$, $\bm{a}^{(i)}$)\
			\State $\Rcal^{(i)} = -\bm{r}_d(\rho_{AV}, \rho_{o}) + \bm{r}_b - \bm{r}_p$
		\EndFor
		\State $\nabla_\phi \Jcal(\phi)  = \frac{1}{N} \sum_{i}^N \nabla_\phi \log (\pi_\phi)\Rcal^{(i)} - \lambda\nabla_\phi\Hcal(\pi_{\phi})$
		\State $\phi = \phi - \alpha\nabla_\phi \Jcal(\phi)$
	\EndFor
\end{algorithmic}
\end{algorithm}

The reason for the simplification is that the orientation of the cyclist only depends on the position of the spawn point given that we want to make the cyclist collide with the AV.
For the trigger distance $D$, it is a conditional probability that depends on the value of the position and the orientation of the spawn point. 
We only adopt one reasonable representation according to prior knowledge, though there could be other reasonable ways.

In this paper, we focus more on critical scenarios that have few participants, e.g., one AV and one dynamic obstacle. However, complex scenarios in the real-world may involve hundreds of vehicles and pedestrians, which is challenging to represent with this method. In those cases, we may need to divide the nodes into groups and model the scenario with groups rather than nodes. This is still an unsolved problem and needs to be studied in the future.

In our implementation, we use Gaussian distribution $\Ncal(\mu, \sigma)$ to model the continuous blocks and multinomial distribution to model the discrete ones. Neural networks (NN) are used for conditional probabilities inference. The reparameterization trick \cite{21} is used to conduct sampling with back-propagation:
\begin{equation}
	\mu_k, \sigma_k \leftarrow \Mcal_k(S, a_{k-1})
\end{equation}
\begin{equation}
	\epsilon \sim \Ncal(0, 1)
\end{equation}
\begin{equation}
	a_k = \mu_k + \sigma_k \times \epsilon
\end{equation}
where $a_k$ is the action sampled from the $k$-th node. $\Mcal_k$ is the model that represents the conditional distribution of the $k$-th action.
Then $a_k$ needs to be rescaled and shifted to represent the parameter of the real-world scenario:
\begin{equation}
	b_k = a_k\times l_k + s_k
\label{rescale}
\end{equation}
where $l_k$ and $s_k$ are the range and average value of the $k$-th action, respectively. We note that $b_k$ will be truncated if its value is beyond the range boundary.

\subsection{Scenario generation framework}

With the language of reinforcement learning, we regard the aforementioned scenario generation model as an agent and the full-stack AV algorithm to be evaluated as an environment. 
Our target is to obtain a safety-critical scenario generation model.

The state of the environment has two parts. 
The first part contains information of route $\bm{\eta}$, which is the reference trajectory for the task algorithm, and the road map with lane information. 
The second part consists of several parameters of the task algorithm $\bm{\xi}$, which influence AV's decision-making ability. 
One simple example is the target speed which our generative model needs to consider to adjust the position of the obstacles.

The reward function consists of three parts:
\begin{equation}
	\Rcal = -\bm{r}_d(\rho_{AV}, \rho_{o}) + \bm{r}_b - \bm{r}_p(\rho_{o}, \gamma)
\end{equation}
where $\rho_{AV}$ and $\rho_{o}$ represents the positions of the AV and the obstacle (the cyclist in our experiments), respectively. The first term is the risk metric $\bm{r}_d$, which we use the distance between the obstacle we generated and the AV to represent:
\begin{equation}
	\bm{r}_d(\rho_{AV}, \rho_{o}) = \| \rho_{AV}-\rho_{o}\|_2
\end{equation}
In the simulation, the distance between two rigid objects is always larger than 0, which means the distance cannot be used to represent the collisions. Therefore, we will provide the agent an extra bonus $\bm{r}_b$ if a collision happens:
\begin{equation}
{\boldsymbol r}_b = 
\left\{
\begin{aligned}
R_b,\ & \text{collision}= True \\
0,\ & \text{collision} = False\\
\end{aligned}
\right.
\end{equation}

\begin{table}[t]
\begin{center}
\caption{Hyper-parameters in our experiments}
\label{parameters}
		\begin{tabular}{c|c|c}
		\hline
		Hyper-parameter    & Description  &  Value \\
		\hline
		$E$  & max epoch number  &  100 \\
		$\alpha$  & learning rate  &  0.008 \\
		$N$  & batch size  &  16 \\
		$\lambda$  & weight for policy entropy  &  0.001 \\
		\hline
		$R_b$  & reward of collision bonus  & 10 \\
		$R_p$  & penalty of route occupy  & 20 \\
		$\gamma$  & threshold in $\bm{r}_p$  & 3 \\
		\hline
		$s_X$  & average value of $X$ & 0 \\
		$s_Y$  & average value of $Y$ & 0 \\
		$s_\Theta$  & average value of $\Theta$ & 180 \\
		$s_D$  & average value of $D$ & 20 \\
		$l_X$  & scale of action $X$ & 100 \\
		$l_Y$  & scale of action $Y$ & 18 \\
		$l_\Theta$  & scale of action $\Theta$ & 360 \\
		$l_D$  & scale of action $D$ & 40 \\
		\hline
		$h_{s}$  & \# of variable in state module  &  64 \\
		$h_{a}$  & \# of variable in action module  &  32 \\
		\hline
		\end{tabular}
\end{center}
\end{table}

Finally, we use a penalty $\bm{r}_p$ to avoid a special case that the obstacles are spawned too close to the route, which is not a reasonable scenario. We use a threshold $gamma$ to determine whether this penalty is implemented:
\begin{equation}
{\boldsymbol r}_p = 
\left\{
\begin{aligned}
R_p,\ & \|\eta_i-\rho_o\|_2 < \gamma\ \forall i \\
0,\ & \text{otherwise} \\
\end{aligned}
\right.
\end{equation}
where $\eta_t$ is the $i$-th route waypoint.

\begin{figure*}[t]
\centering
\includegraphics[width=17.6cm]{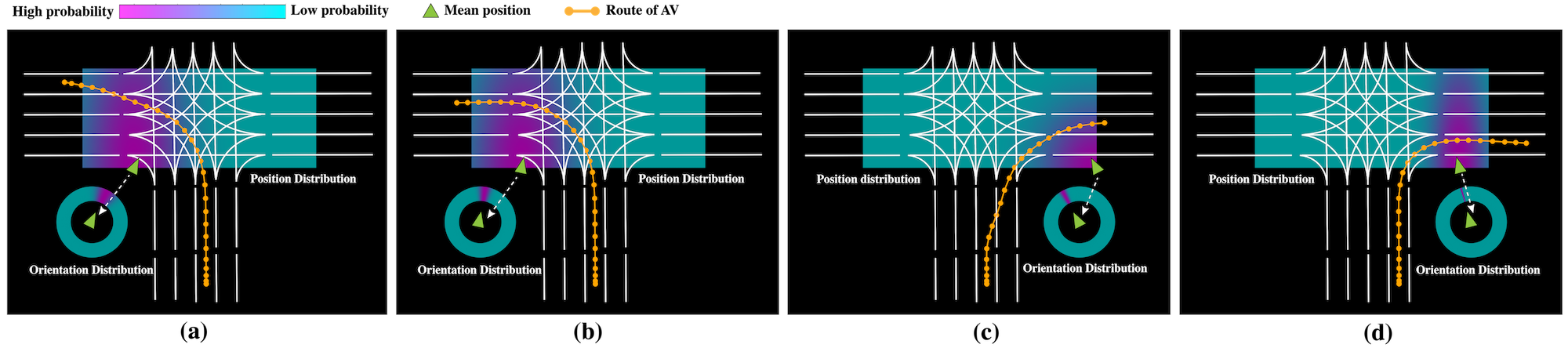}
\caption{The probabilities of $P(X|S)$, $P(Y|S)$ and $P(\Theta|X,Y,S)$ are given with different colors, where the pink is high probability and blue is low probability. The rectangle represents the probabilities of position X $P(X|S)$ and position Y $P(Y|S)$. The circle represents the probability of orientation $P(\Theta|X=x,Y=y,S=s)$ when the position X and Y are conditioned on the mean values of $P(X=x|S=s)$ and $P(Y=y|S=s)$.}
\label{action_space}
\end{figure*}

\subsection{Optimization process}

We follow the policy gradient method REINFORCE \cite{25} to solve our optimization problem. The gradient for updating model parameter $\phi$ is:
\begin{equation}
\begin{split}
	\nabla_\phi \Jcal(\phi) 
	=& E_{a\sim \pi_\phi}\left[ \nabla_\phi \log (\pi_\phi)  \right] \Rcal(a) \\
	\approx & \frac{1}{N} \sum_{i}^N \nabla_\phi \log (\pi_\phi(a_i))\Rcal(a_i)
\end{split}
\end{equation} 
where $\Jcal(\phi) = E_{a\sim \pi_\phi} [\Rcal]$ is the objective function, and $a$ is the action sampled from the policy distribution $\pi_\phi$. To encourage the diversity of the policy, we add an entropy term to the objective function as proposed in \cite{31}:
\begin{equation}
	H(\pi_\phi) = -\int \pi_\phi(x)\log \pi_\phi(x)dx
\end{equation}

When we use an autoregressive Gaussian distribution to model the policy $\pi_\phi$, we are still able to calculate the joint probability with chain rule:
\begin{equation}
	\log P(\bm\pi) = \log P(\pi_0) + \sum_{i=1}^n \log P(\pi_i|\pi_0,...,\pi_{i-1})
\end{equation}
and each term is calculated according to the density function of Gaussian distribution. The entropy is also easy to calculate since the joint distribution is still a Gaussian distribution. The entire algorithm is shown in \textbf{Algorithm}~\ref{alg1}.

\begin{figure}[t]
\centering
\includegraphics[width=7cm]{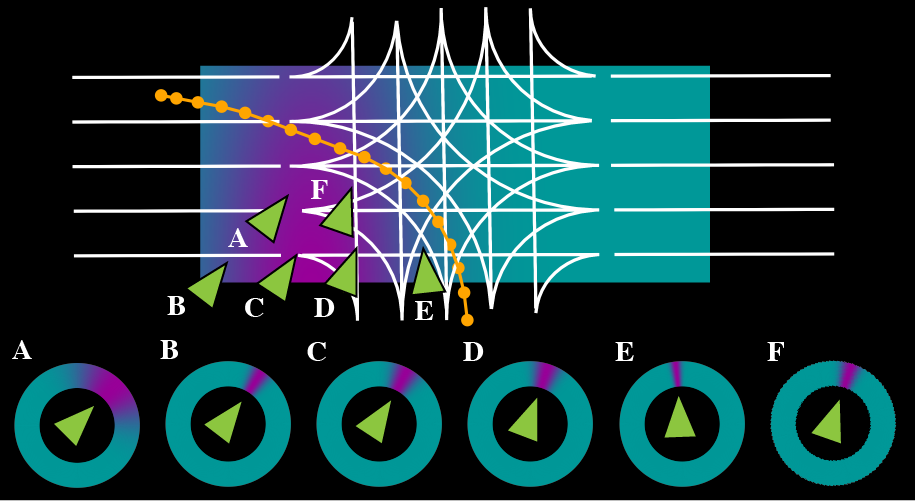}
\caption{Six positions are sampled and considered as conditions to output $P(\Theta|X=x, Y=y, S=s)$. The conditional probabilities are shown at the bottom with their probability density functions (pdf).}
\label{autoregressive}
\end{figure}

\begin{figure}[t]
\centering
\includegraphics[width=8.6cm]{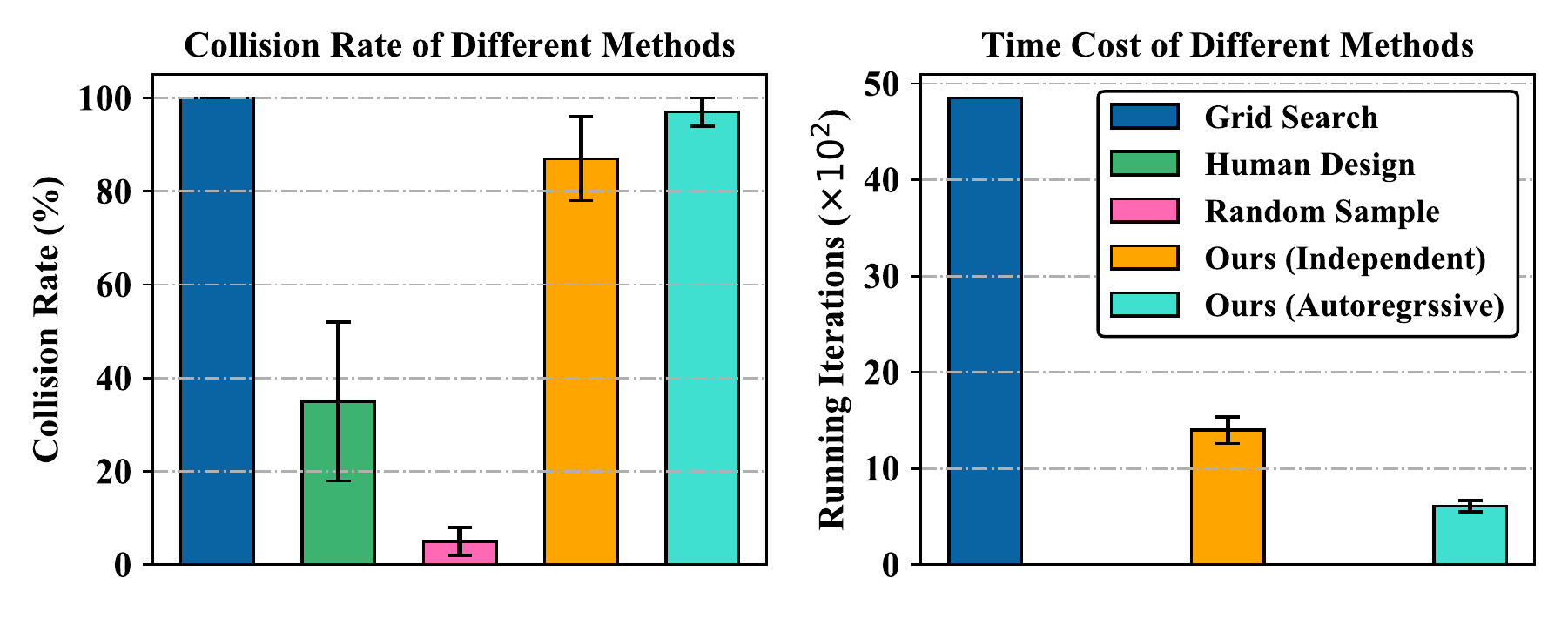}
\caption{The comparative results of the collision rate and the number of iterations required to reach stability. Only three methods are compared with the second metric since the other two methods do not require iteration.}
\label{compare}
\end{figure}

\section{Experiment}

\subsection{Experimental settings}

We implemented our algorithm with Pytorch \cite{33} and use Adam \cite{30} as the optimizer. 
Details about the hyper-parameter of our experiments are listed in Table.~\ref{parameters}. 
We use \textit{Carla} \cite{24} as our simulation platform, and we use the Carla Scenario Runner Library \cite{32} as the backbone to generate traffic scenarios. 
We modify the \textit{Scenario04} and use it as our test-bed to evaluate our framework. 
The description of the setting of this scenario is discussed in Section~\ref{scenario04}. 

The AV algorithm to be evaluated is a simple trajectory following model with a PID controller. 
we use the visualization and debugging tool \textit{Carla-display}, which is an open-source tool developed by our group \cite{34}. 
We use a single fully-connected layer to build our state and action modules, and the number of hidden variables is summarized in Table.~\ref{parameters}.

\subsection{Verification experiment}

We conduct a verification experiment to show that our proposed framework can generate risky scenarios. 
We train our model on 10 different routes and randomly sample target speeds from $20km/h$ to $50km/h$. 
Then we test this model on 4 different routes and the results are displayed in Fig.~\ref{action_space}. 
The pink color represents high probability output of the policy that will result in high risk scenarios.
It is shown that our model outputs different policies when different routes and speeds are fed in. 
For left-turn scenarios, the distributions of the positions fall in the left of the intersection; and for right-turn scenarios, the distributions are on the right. 

We note that the high probability regions are exactly the most dangerous ones a human driver will encounter in reality, i.e., the driver tends to ignore these blind spots when passing through these intersections.

To verify the contribution of our autoregressive structure, we sample different initial positions ($x$ and $y$) from $P(X|S)$ and $P(Y|S)$ and use them as the conditions to obtain the orientation output $P(\Theta|X=x, Y=y, S=s)$. 
The results are shown in Fig.~\ref{autoregressive}. As we expected, when different $x$ and $y$ are used as conditions, the orientation is quite different, which proves that our model can learn the dependencies between different policies.

\begin{figure*}[t]
\centering
\includegraphics[width=17.5cm]{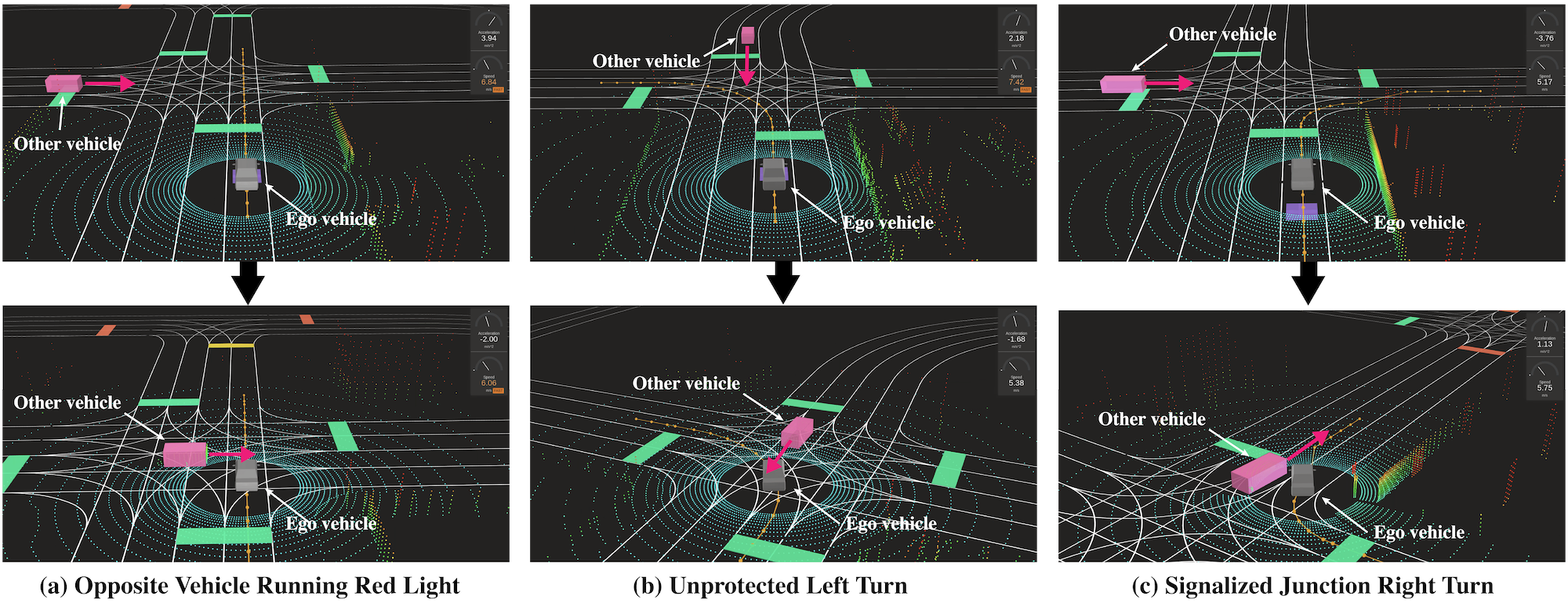}
\caption{Generated risky scenarios for three new settings. The first row shows the initial state and the second row shows the pre-crash state. All of these scenario settings are implemented with the Carla Scenario Runner Library \cite{32}.}
\label{more}
\end{figure*}

\subsection{Comparison experiment}

In this section, we construct four baselines for comparison:
\subsubsection{Grid Search} This is the easiest way to search the risk scenarios. Since we need to consider the combinations of all policies, the search space could exponentially grow up as the dimension of scenario representation increases. To reduce the searching expense, we discretize the policies by steps [4, 3, 20, 10] for [$X$, $Y$, $\Theta$, $D$] and search all combinations to solutions.

\subsubsection{Human Design} The Carla Scenario Runner Library \cite{32} is used for artificially creating scenarios in Carla Challenge \footnote{\url{https://carlachallenge.org/}}. This competition aims at testing and evaluating AV algorithms in risky scenarios. We use the same parameters and procedures of the testbed of this competition as our baseline.

\subsubsection{Random Sampling} In this baseline, We sample all policies from the uniform distribution. This method is designed to simulate the methods that are not combined with task algorithms, in which case only random scenarios are generated.

\subsubsection{Independent Policy} This method is almost the same as our proposed one. The only difference is we treat all policies as independent blocks, i.e., each policy is modeled by a Gaussian distribution only conditioned on the state $S$. 

We use two metrics for comparison: the collision rate after the model achieves a stable policy and the number of iterations required for the model to reach stability. We conduct 30 experiments on each method with different routes and target speeds. The experimental results are shown in Fig.~\ref{compare}. 

The results of the collision rate show that the human design method has very poor adaptability (large variance) because the artificially designed parameters are only useful when the AV algorithm satisfies some conditions, e.g, the target speed belongs to a specific region. 
For the random sampling method, the low rate is in line with our expectations since there is no guidance to help generate risky scenarios. 
The independent method has a slightly lower collision rate than our method because the independence among the actions makes the generating process very inefficient. Decoupling the policies may lose some constraints.

The results of the time cost show that the \textit{Independent Policy} method requires more time to reach the stable state than our method because the variance of each policy could influence other policies, which makes the training process unstable. 
Even if we have selected large steps to discretize the space (leading to very few collision scenarios), the grid search method still requires too much time. 

In Fig.~\ref{compare}, all results have a variance with the exception of the grid search, because, for different routes or target speeds, the search process needs to be done from scratch, which is a tremendous drawback of the grid search method. All of the other methods have the adaptiveness to generate different risky scenarios according to the input states. The lower the variance is, the more adaptive the method is.

\subsection{Experiments on other scenarios}

We test three other scenarios (also appear in \cite{32}) to verify the effectiveness of our method. We choose these scenarios because all of them cause millions of losses every year according to the pre-crash scenarios report from the National Highway Traffic Safety Administration (NHTSA) \cite{40}.

\subsubsection{Opposite Vehicle Running Red Light} 
The ego vehicle passes through an intersection along with a straight route, while another vehicle takes the priority from the ego vehicle by running a red traffic light. 
The action space has three dimensions: the position $X$, the position $Y$ and the speed $V$. 
The orientation of the other vehicle is fixed in this scenario.

\subsubsection{Unprotected Left Turn} 
The ego vehicle turns left, while another vehicle approaches from the opposite direction. 
To avoid a collision, the ego vehicle should wait for the other vehicle or speed up to cross the intersection. The action space has the same three dimensions as the last scenario.

\subsubsection{Signalized Junction Right Turn} 
The ego vehicle turns right, while another vehicle approaches from the left. 
The action space has the same three dimensions as the last scenario.

The generated risky scenarios with our framework are shown in Fig.~\ref{more}. 
As displayed in the figure, in all three new scenarios, our method also finds the risky distributions of all building blocks.

\subsection{Exploration of solution space}

\begin{figure}[t]
\centering
\includegraphics[width=8cm]{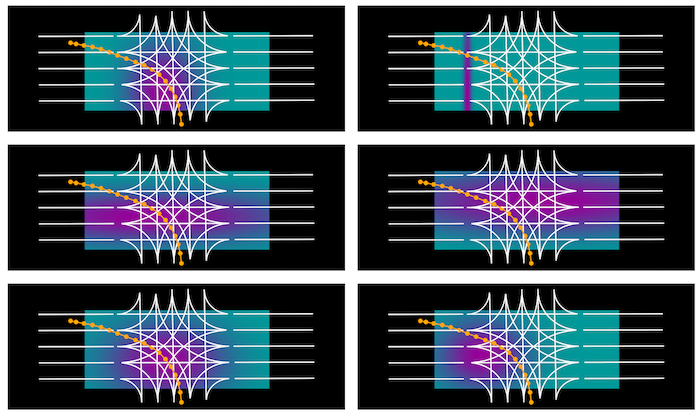}
\caption{The training results have large diversity since the feasible solution is a continuous space rather than a single point. Six different stable policies are shown in this figure.}
\label{diversity}
\end{figure}

We explore the solution space of the risky scenario parameters. Six different stable solutions during our experiments are shown in Fig.~\ref{diversity}. All of them are initialized with different values. 
Although we model the solution with Gaussian distributions, the solution space may not be a convex set, even not only have one mode. For example, both sides of the route should be feasible solutions, but using REINFORCE algorithm with Gaussian policies only models a sub-space of the entire solution space.

In our future work, one possible improvement is separating the action space and the scenario representation space. While the action space is still modeled by Gaussian distribution, the scenarios are modeled with implicit and complex distributions. Tools such as VAE \cite{21} and flow-based models \cite{22} could be leveraged to build a mapping from a simple Gaussian distribution to a complex distribution. Then a joint optimization of the mapping model and the generative model should be done to reach our goal.

\section{Conclusion}

This paper proposes an adaptive framework for safety-critical scenario generation. Firstly, we divide the traffic scenarios into some reusable building blocks and build dependency among them with human knowledge. With this representation, we can generate scenarios based on probability distributions. Then, with the idea borrowed from reinforcement learning, we combine the specific task algorithm with the aforementioned generative model to generate risky scenarios.

Experimental results verify that the proposed framework can generate risky scenes in line with daily experience, and also show that our method has two advantages compared with baselines: (1) the efficiency of our algorithm is much higher than grid search and human design; (2) our generation process is based on the sampling of the probability conditioned on the task algorithms, which has stronger adaptability and diversity.

Future work will focus on building a mapping from the action space to scenario representation space to model the multimodal distribution of safety-critical scenarios. The combination of our evaluation method and the task algorithm under the adversarial training framework could also provide a promising way to develop more robust agents. We will also explore a scalable way to extend our representation method to more complex scenarios.

\section*{Acknowledgment}
This research was sponsored in part by Bosch. The authors would like to thank the Bosch team for their valuable discussion.

\bibliographystyle{IEEEtran}
\bibliography{main}

\end{document}